\documentclass[letterpaper]{article} 
\usepackage[preprint]{aaai2027}  
\usepackage[hyphens]{url}  
\usepackage{graphicx} 
\usepackage{natbib}  
\usepackage{caption} 
\usepackage{amsmath}
\usepackage{amssymb}
\usepackage{booktabs}
\usepackage{multirow}

\newcommand{\mat}[1]{\mathbf{#1}}
\newcommand{\vect}[1]{\boldsymbol{#1}}
\newcommand{\mcal}[1]{\mathcal{#1}}
\newcommand{\tok}[1]{\langle\text{\texttt{#1}}\rangle}

\title{GLaQ: Grounding Latent Queries in Visual Evidence for Multimodal Reasoning}
\author{
Zesheng Yang\textsuperscript{1},
Lingling Zhang\textsuperscript{1}\corresponding,
Xinyu Zhang\textsuperscript{2},
Cheng Zhang\textsuperscript{1},\\
Pengyu Li\textsuperscript{1},
Heng Wang\textsuperscript{1},
Lin Wu\textsuperscript{1}
}
\affiliations{
\textsuperscript{1}Xi'an Jiaotong University\\
\textsuperscript{2}Xidian University
}

\begin{document}

\maketitle

\begin{abstract}
Chain-of-thought reasoning has substantially improved the problem-solving capabilities of multimodal large language models. Fine-grained visual evidence, however, remains difficult to preserve and reuse across text-based reasoning steps. To address this limitation, tool-augmented thinking-with-images methods maintain visual access externally by revisiting or manipulating the image, but require predefined tools and additional inference-time processing. As an internal alternative, continuous visual latent reasoning retains intermediate computation in hidden states. However, its prevailing autoregressive construction makes each latent state depend on its predecessors, so later states may repeat information already present in the latent sequence rather than capture complementary visual details. We introduce GLaQ, a grounded latent-query framework that replaces sequential latent rollout with a fixed set of context-conditioned queries grounded in the original visual tokens. The grounded queries are reinjected for answer generation, providing direct and coordinated access to source visual evidence. We train GLaQ with localized-view supervision followed by reinforcement learning under task-level rewards. Across five benchmarks for fine-grained visual understanding and perception, GLaQ-7B gains 5.99--9.66\% over its base model and leads all compared visual latent methods, suggesting that direct query-to-image grounding can recover localized evidence from the full image without external visual operations or autoregressive latent rollouts.
\end{abstract}

\section{Introduction}

\begin{figure}[!t]
    \centering
    \includegraphics[width=\columnwidth]{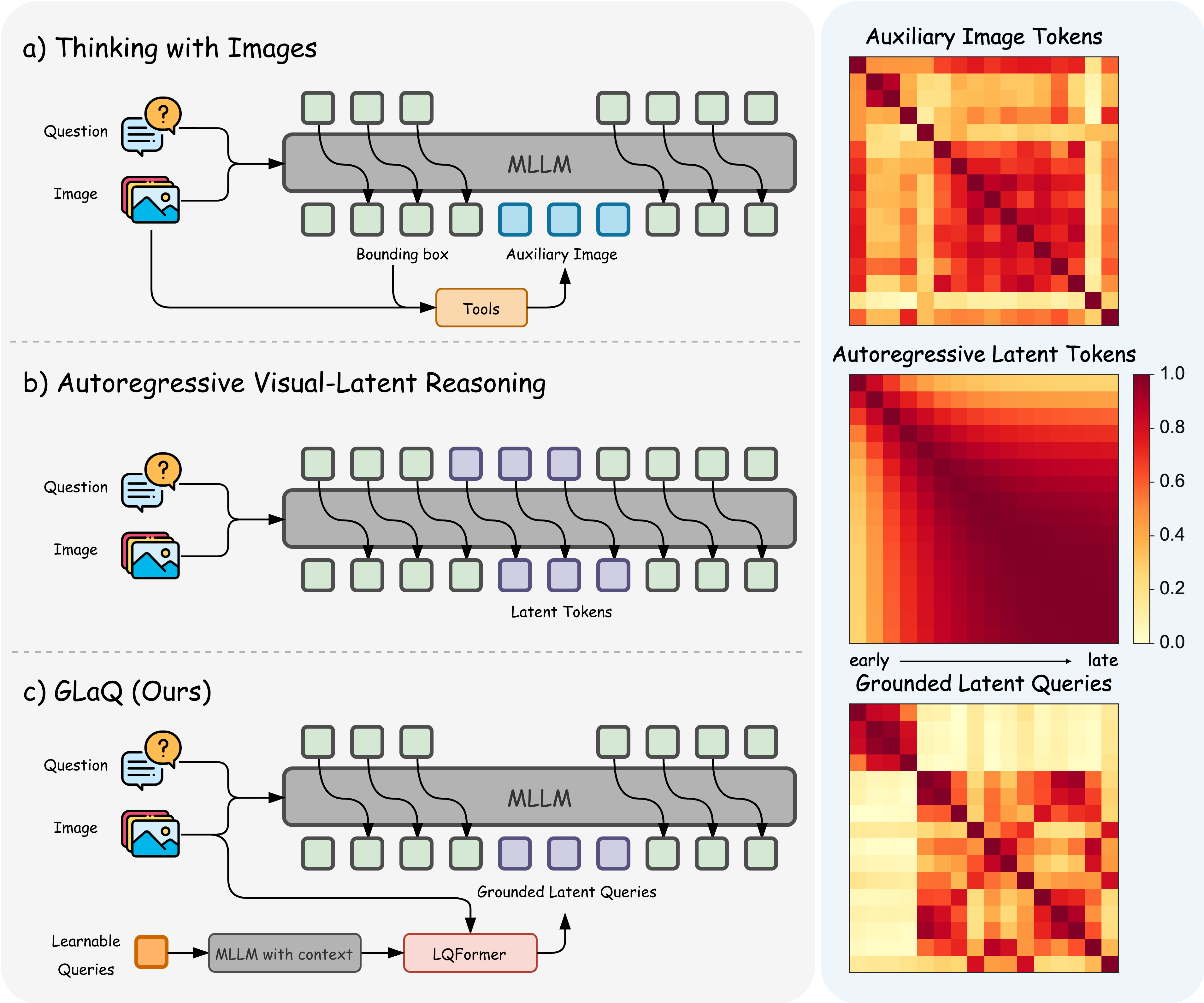}
    \caption{Comparison of visual reasoning paradigms. We compare (a) thinking with images, (b) autoregressive visual latent reasoning, and (c) GLaQ. The heatmaps show within-set pairwise cosine similarities for the three representations; darker indicates higher similarity.}
    \label{fig:intro-comparison}
    \vspace{-8pt}
\end{figure}

Multimodal large language models (MLLMs) benefit from chain-of-thought reasoning, achieving strong performance across diverse visual reasoning tasks \citep{zhang2023multimodalcot}. However, standard multimodal chain-of-thought verbalizes intermediate decisions as image-conditioned textual rationales, creating an information bottleneck in which fine-grained visual details may be compressed or lost before reuse. To overcome this bottleneck, thinking-with-images methods revisit or manipulate the image by localizing relevant regions and applying operations such as cropping, zooming, or drawing annotations \citep{hu2024visualsketchpad,wu2024vstar,shen2025zoomeye,zheng2026deepeyes}, supplying subsequent reasoning steps with refreshed visual evidence. While effective, these methods require predefined tools and additional inference-time processing. Continuous visual latent reasoning instead retains intermediate computation in hidden states \citep{hao2024coconut,li2025lvr,wang2025monet,cheng2026hylar}, providing a compact internal alternative that carries visual semantics across reasoning steps without text conversion or external image operations.

However, prevailing visual latent methods construct their latent sequences autoregressively. Each latent state depends on its predecessors, mirroring next-token decoding \citep{li2025lvr,tong2025skila,wang2025monet,cheng2026hylar}. This ordering suits language generation, whereas visual latent states collectively form an internal inference space. The similarity maps in Figure~\ref{fig:intro-comparison} expose this limitation: auxiliary image tokens exhibit heterogeneous pairwise relations, whereas autoregressive latent tokens become increasingly alike, forming a high-similarity block late in the sequence. This convergence suggests that later states may repeat shared content rather than add complementary visual evidence. The resulting challenge is to retain the compactness of latent reasoning while grounding each latent state directly in the source visual representation and coordinating the states without an autoregressive dependency chain.

To address this challenge, we propose \textbf{G}rounded \textbf{La}tent \textbf{Q}ueries (\textbf{GLaQ}), a framework that replaces autoregressive latent rollout with a fixed block of context-conditioned queries grounded in the original visual tokens, as illustrated in Figure~\ref{fig:intro-comparison}(c). GLaQ follows a \emph{contextualize--ground--reinject} process. The current multimodal context first shapes the query block, which a lightweight Latent Query Former (LQFormer) grounds in the original visual tokens before the resulting queries are reinjected as soft visual prompts for answer generation. GLaQ's grounded queries instead show a heterogeneous pairwise structure qualitatively closer to auxiliary image tokens than to later autoregressive states, suggesting that coordinated query slots retain differentiated visual information without external image operations or a sequential latent chain. To train this interface, self-distillation from region-of-interest (ROI) views transfers localized evidence to the full-image inference path, while Decoupled Policy Optimization (DePO) \citep{cheng2026hylar} optimizes both textual and grounded-query actions using task-level rewards.

Across five benchmarks spanning fine-grained perception, high-resolution understanding, and real-world reasoning, GLaQ-7B outperforms its Qwen2.5-VL-7B base model \citep{bai2025qwen25vl} by 5.99--9.66\% and leads all compared visual latent methods on every benchmark. These results suggest that direct query-to-image grounding offers an effective way to recover localized evidence from the full image without external visual operations.

Our main contributions are summarized as follows:
\begin{itemize}
    \item We introduce GLaQ, a grounded latent-query interface that replaces autoregressive latent rollout with a fixed context-conditioned query block. Its \emph{contextualize--ground--reinject} pathway provides direct and coordinated access to source visual tokens.
    \item We develop a learning strategy for grounded latent queries that combines supervised and reinforcement learning. ROI-guided self-distillation transfers localized evidence to the full-image inference path, while DePO directly optimizes the grounded queries under task-level rewards.
    \item Across five benchmarks, GLaQ-7B outperforms its base model and ranks first among all compared visual latent methods, demonstrating the effectiveness of grounded latent queries for visual reasoning.
\end{itemize}

\section{Related Work}

\paragraph{Explicit Multimodal Reasoning.}
Explicit multimodal reasoning externalizes computation as text or image-space operations. Multimodal-CoT and LLaVA-CoT produce image-conditioned textual rationales \citep{zhang2023multimodalcot,xu2025llavacot}. R1-Onevision verbalizes visual content; Vision-R1 and VL-Rethinker refine rationales through outcome-driven learning and self-reflection \citep{yang2025r1onevision,huang2026visionr1,wang2025vlrethinker}. Although interpretable, textual traces must verbalize fine-grained attributes and spatial relations before reuse. Thinking-with-images methods instead revisit or manipulate images: Visual CoT, V$^\ast$, and ZoomEye localize and revisit question-relevant regions for finer-grained evidence \citep{shao2024visualcot,wu2024vstar,shen2025zoomeye}. DeepEyes and ToolsRL use reinforcement learning to adaptively select and invoke visual tools \citep{zheng2026deepeyes,dong2026toolsrl}. Visual Sketchpad, Thyme, CodeVision, and CodeDance execute programs that compose image operations and expose intermediate results \citep{hu2024visualsketchpad,zhang2025thyme,guo2026codevision,song2026codedance}. These methods expose hard-to-verbalize evidence but require external execution and additional image processing and remain limited to supported operations.

\paragraph{Continuous Visual Latent Reasoning.}
Continuous latent reasoning retains intermediate computation in hidden states rather than discrete language. Coconut, CODI, and SemCoT develop recurrent, self-distilled, or semantically aligned language-only latent reasoning \citep{hao2024coconut,shen2025codi,he2025semcot}. The paradigm then extends to vision: Mirage, LVR, and SkiLa distill or reconstruct visual states \citep{yang2025mirage,li2025lvr,tong2025skila}; ILVR and VaLR integrate evolving or vision-aligned states \citep{dong2025ilvr,jeon2026valr}. Monet, HyLaR, LanteRn, and DeepSketcher learn interleaved text--latent trajectories \citep{wang2025monet,cheng2026hylar,viveiros2026lantern,zhang2026deepsketcher}; Latent Implicit Visual Reasoning and Mull-Tokens use task-adaptive or modality-agnostic tokens \citep{li2026latentimplicit,ray2026mulltokens}. Recent work improves grounding through teacher alignment, spatial-semantic supervision, region-level attributes, relation modeling, and input-compatible feature alignment \citep{wu2026lavit,cui2026ris,xu2026slvr,wang2026regular,miao2026gap}. Yet many visual latent methods generate states autoregressively, so later representations depend on earlier ones; recent analyses identify information-gain collapse, reconstruction mismatch, and latent bypass \citep{wang2026scolar,adhikari2026pearl,viveiros2026holdingback}, suggesting longer trajectories need not contribute complementary visual information.

\section{Method}

\begin{figure*}[!t]
\centering
\includegraphics[width=\textwidth]{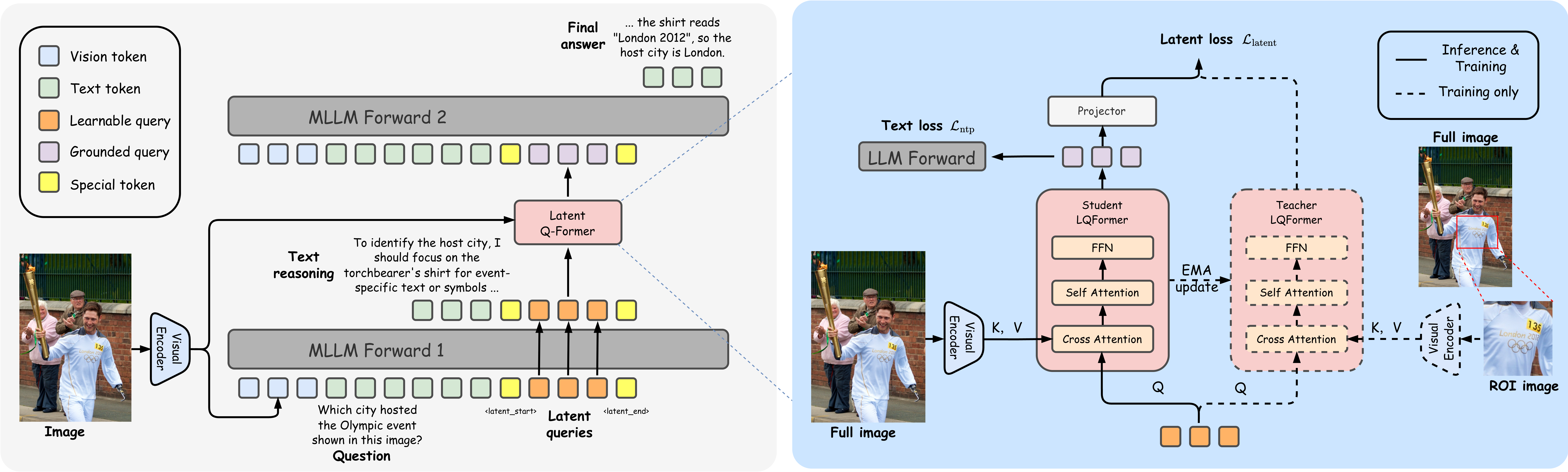}
\caption{Overview of GLaQ. \textit{Left:} The first pass through a multimodal large language model (MLLM) contextualizes a fixed query block under the current image--text prefix. Latent Query Former (LQFormer) grounds the resulting queries in source visual tokens, and the grounded block is reinjected into the second pass as soft prompts for answer generation. \textit{Right:} During supervised fine-tuning, a full-image student and a region-of-interest (ROI) view teacher updated by an exponential moving average (EMA) ground the same contextualized queries. Slot-wise latent alignment transfers localized evidence to the student, while next-token prediction trains response generation; the teacher is removed at inference.}
\label{fig:glq-method}
\vspace{-8pt}
\end{figure*}

\subsection{Method Overview}

GLaQ introduces grounded latent queries as a soft visual prompting mechanism for multimodal reasoning. Rather than generating a latent segment autoregressively \citep{li2025lvr,tong2025skila,wang2025monet,cheng2026hylar}, GLaQ instantiates a fixed block of query slots and grounds their contextualized states directly in the original visual tokens. This design preserves continuous intermediate computation while allowing the queries to access and coordinate visual information without forming a sequential latent chain.

As illustrated in Figure~\ref{fig:glq-method}, GLaQ follows a two-pass \emph{contextualize--ground--reinject} process. When the MLLM emits a latent-start token, a fixed block of learnable query slots is inserted into the decoding stream. In the first pass, the MLLM contextualizes these slots using the current multimodal prefix, after which a lightweight Latent Query Former (LQFormer) grounds the resulting states in the original visual tokens and coordinates information across query slots. In the second pass, the grounded queries replace the initial query slots and are reinserted as soft visual prompts, after which the model resumes standard autoregressive text generation.

GLaQ is trained through supervised latent-query alignment followed by reinforcement learning. During supervised training, a teacher grounded on a region-of-interest (ROI) view provides localized targets for a student operating on the full image, transferring localized supervision to the inference-time grounding pathway. GLaQ is then refined with DePO, which optimizes categorical text tokens and continuous grounded-query actions under task-level rewards.

\subsection{Grounded Latent Query}
\label{sec:grounded-latent-queries}

Grounded Latent Query inserts a continuous visual-query segment into the decoding stream of an MLLM. Unlike autoregressive latent-token methods that generate latent states one by one, GLaQ instantiates the entire $K$-slot query block jointly. The inserted slots are first contextualized by the current language-image prefix and then grounded against the original visual tokens. The grounded slots are reinserted as soft prompts, so subsequent text decoding can use retrieved visual evidence while retaining standard autoregressive decoding. The left panel of Figure~\ref{fig:glq-method} illustrates this inference-time latent-query pathway.

Given an input image $\mat{I}$ and a text prefix $\vect{x}$, a frozen visual encoder maps the image into visual tokens $\mat{V}=f_v(\mat{I})=[\vect{v}_1;\ldots;\vect{v}_N]\in\mathbb{R}^{N\times d}$, where $\vect{v}_i\in\mathbb{R}^d$.
The multimodal language model $\mcal{M}_\theta$ decodes from $(\vect{x},\mat{V})$ until it emits $\tok{latent\_start}$. The resulting latent segment contains the full query block $\mat{Q}=[\vect{q}_1;\ldots;\vect{q}_K]\in\mathbb{R}^{K\times d}$ followed by a closing marker $\tok{latent\_end}$:
\begin{equation}
\begin{gathered}
\tok{latent\_start},\;
\underbrace{\vect{q}_1,\ldots,\vect{q}_K}_{K\ \mathrm{query\ slots}},\;
\tok{latent\_end} .
\end{gathered}
\end{equation}
The query slots are learnable continuous embeddings optimized jointly with the model, whereas the boundary markers remain ordinary text targets. Inserting all $K$ slots jointly contextualizes them in one forward pass, avoiding $K$ autoregressive latent-generation steps and preserving stable slot identities for alignment.

\subsubsection{Two-Pass Latent Forward}

GLaQ resolves each latent segment in two passes. The first pass contextualizes the query slots under the current language-image prefix into $\mat{U}_Q=[\vect{u}_1;\ldots;\vect{u}_K]$, which LQFormer $\mcal{F}_\phi$ grounds against $\mat{V}$ to obtain $\mat{Z}_Q=[\vect{z}_1;\ldots;\vect{z}_K]$. The second pass replaces $\mat{Q}$ with $\mat{Z}_Q$ and resumes text decoding:
\begin{equation}
\begin{aligned}
\mat{U}_{Q}
&=
\mcal{M}^{(1)}_\theta(\vect{x},\mat{V};\mat{Q}),\\
\mat{Z}_{Q}
&=
\mcal{F}_\phi(\mat{U}_{Q},\mat{V}),\\
p_\theta(y_t \mid \vect{y}_{<t},\vect{x},\mat{V},\mat{Z}_{Q})
&=
\mcal{M}^{(2)}_\theta(y_t \mid \vect{y}_{<t},\vect{x},\mat{V};\mat{Z}_{Q}).
\end{aligned}
\end{equation}
Thus, GLaQ implements a retrieve-and-insert process: the first pass forms visual evidence requests, LQFormer grounds them in the image, and the second pass uses the grounded queries as soft prompts for answer generation.
When a sequence contains multiple grounded-query blocks, the same two-pass operation is applied at each block using the prefix available at that position.

\subsubsection{Latent Query Former}

LQFormer is the grounding module that converts context-conditioned evidence requests into image-grounded query prompts. Starting from $\mat{U}^{0}=\mat{U}_{Q}$, it stacks $L$ lightweight query-former blocks, following the Q-Former design for bridging frozen visual representations and language models \citep{li2023blip2}. Our main model uses $L=2$ and outputs $\mat{Z}_{Q}=\operatorname{RMSNorm}(\mat{U}^{L})$. For each layer $\ell\in\{0,\ldots,L-1\}$, cross-attention first lets the latent slots retrieve image evidence, self-attention then lets the slots exchange complementary evidence, and a feed-forward network (FFN) refines the grounded states:
\begin{equation}
\begin{aligned}
\bar{\mat{U}}^{\ell}
&=
\mat{U}^{\ell}
+
\operatorname{CrossAttn}^{\ell}
\left(
\operatorname{RMSNorm}(\mat{U}^{\ell}),\mat{V}
\right),\\
\tilde{\mat{U}}^{\ell}
&=
\bar{\mat{U}}^{\ell}
+
\operatorname{SelfAttn}^{\ell}
\left(
\operatorname{RMSNorm}(\bar{\mat{U}}^{\ell})
\right),\\
\mat{U}^{\ell+1}
&=
\tilde{\mat{U}}^{\ell}
+
\operatorname{FFN}^{\ell}
\left(
\operatorname{RMSNorm}(\tilde{\mat{U}}^{\ell})
\right).
\end{aligned}
\end{equation}
In cross-attention, the contextualized query states serve as queries and $\mat{V}$ provides keys and values, while self-attention coordinates the $K$ slots. Because only the query slots interact with the $N$ visual tokens, LQFormer localizes the new retrieval behavior in a lightweight adapter without modifying the visual encoder or MLLM decoder.

\subsection{Supervised Fine-Tuning with Latent Query Alignment}
\label{sec:learning-glq}

The inference pathway in Section~\ref{sec:grounded-latent-queries} gives each query direct access to the source visual tokens, but does not by itself specify which evidence should be retrieved. To guide query grounding, we use auxiliary visual views that make task-relevant evidence more salient during SFT. Because inference receives only the original image, we transfer this localized supervision to full-image queries through latent-query alignment, jointly optimized with next-token prediction:
\begin{equation}
\mathcal{L}
=
\lambda_{\mathrm{latent}}\mathcal{L}_{\mathrm{latent}}
+ \mathcal{L}_{\mathrm{ntp}},
\end{equation}
where $\mathcal{L}_{\mathrm{latent}}$ transfers localized visual supervision to full-image grounded queries, and $\mathcal{L}_{\mathrm{ntp}}$ teaches the model to use these queries for response generation. We set $\lambda_{\mathrm{latent}}=0.3$ throughout SFT.

\subsubsection{Latent Query Self-Distillation}

For each query block, the student LQFormer grounds the contextualized queries $\mat{U}_Q$ in the original image, while an exponential moving average (EMA) teacher grounds the same queries in the paired auxiliary view $\mat{I}_{\mathrm{roi}}$. Because both paths receive the same $\mat{U}_Q$, their evidence requests and slot ordering remain matched, allowing each auxiliary-view teacher query to supervise its full-image student counterpart:
\begin{equation}
\begin{aligned}
\mat{Z}_{Q}^{\mathrm{stu}}
&= \mcal{F}_\phi(\mat{U}_{Q}, f_v(\mat{I})),\\
\mat{Z}_{Q}^{\mathrm{tea}}
&=
\operatorname{StopGrad}\!\left(
\mcal{F}_{\bar{\phi}}(\mat{U}_{Q}, f_v(\mat{I}_{\mathrm{roi}}))
\right),\\
\bar{\phi}
&\leftarrow \tau\bar{\phi} + (1-\tau)\phi .
\end{aligned}
\end{equation}
The teacher has the same architecture as the student LQFormer but receives no gradients; its parameters $\bar{\phi}$ are updated solely by EMA from the student parameters $\phi$ with momentum $\tau$, following the asymmetric online--target design of BYOL \citep{grill2020byol}. Only the LQFormer is maintained as an EMA teacher; the MLLM backbone is shared by the two paths.

A projector $g_\psi(\cdot)$ maps the student queries into the alignment space. For a response containing $B$ grounded-query blocks, let $\hat{\vect{p}}_{b,k}=\operatorname{norm}(g_\psi(\vect{z}_{b,k}^{\mathrm{stu}}))$ and $\hat{\vect{t}}_{b,k}=\operatorname{norm}(\vect{z}_{b,k}^{\mathrm{tea}})$. Because both paths receive the same ordered query block, their slots correspond directly. We align all $BK$ pairs using:
\begin{equation}
\begin{aligned}
\mathcal{L}_{\mathrm{latent}}
&=
\frac{1}{BK}\sum_{b=1}^{B}\sum_{k=1}^{K}
\left\|
\hat{\vect{p}}_{b,k}-\hat{\vect{t}}_{b,k}
\right\|_2^2\\
&=
\frac{1}{BK}\sum_{b=1}^{B}\sum_{k=1}^{K}
\left[
2-2\cos(\hat{\vect{p}}_{b,k},\hat{\vect{t}}_{b,k})
\right].
\end{aligned}
\end{equation}

The projector confines alignment to a prediction space, leaving the student queries themselves available for second-pass generation. This objective transfers evidence exposed by the auxiliary view to queries grounded in the full image, without reconstructing dense visual tokens. The teacher and auxiliary views are removed at inference.

\subsubsection{Two-Stage Joint Text--Latent Supervision}

Latent alignment determines what evidence the queries should retrieve, but not how the resulting grounded states should support a reasoning trajectory. We therefore apply next-token prediction to the surrounding text and latent boundary markers, while excluding the internal continuous query slots from vocabulary supervision. Let $T$ denote the number of textual target positions in the response, and let $\mat{Z}_{Q,<t}$ denote the concatenation of all grounded-query blocks preceding position $t$. The text objective is:
\begin{equation}
\mathcal{L}_{\mathrm{ntp}}
=
-\frac{1}{T}
\sum_{t=1}^{T}
\log p_\theta(y_t \mid \vect{y}_{<t},\vect{x},\mat{V},\mat{Z}_{Q,<t}).
\end{equation}

Stage 1 uses short-answer Visual-CoT examples \citep{shao2024visualcot}, placing one grounded-query block before the answer and aligning it with the paired localized view. Stage 2 starts from the Stage 1 student checkpoint, with the EMA LQFormer teacher initialized from the student LQFormer and the student-side projector randomly reinitialized. It then trains on Zebra-CoT visual-search trajectories \citep{li2025zebracot}, in which interleaved reasoning views are represented by grounded-query blocks between textual reasoning steps. Training on these trajectories teaches the model when to interleave grounded-query blocks with textual reasoning, providing a supervised cold start for the subsequent reinforcement-learning stage. Across both stages, $\mathcal{L}_{\mathrm{latent}}$ supervises what each block retrieves, while $\mathcal{L}_{\mathrm{ntp}}$ teaches the model to incorporate the retrieved evidence into intermediate reasoning and final prediction.

\begin{table*}[t]
\centering
{\small
\setlength{\tabcolsep}{0.9pt}
\renewcommand{\arraystretch}{1.08}
\begin{tabular*}{\textwidth}{@{\extracolsep{\fill}}lccccccccccccc@{}}
\toprule
\multirow{2}{*}[-0.4ex]{\textbf{Methods}} & \multicolumn{3}{c}{\textbf{V$^\ast$}} & \multicolumn{3}{c}{\textbf{HRBench-4K}} & \multicolumn{3}{c}{\textbf{HRBench-8K}} & \multicolumn{3}{c}{\textbf{MME-RealWorld-Lite}} & \multicolumn{1}{c}{\textbf{MMVP}} \\
\cmidrule(lr){2-4}\cmidrule(lr){5-7}\cmidrule(lr){8-10}\cmidrule(lr){11-13}\cmidrule(l){14-14}
& Overall & Attr. & Spat. & Overall & Single & Cross & Overall & Single & Cross & Overall & Perc. & Reas. & Overall \\
\midrule
\multicolumn{14}{c}{\textbf{\textit{Proprietary Model}}} \\
\cmidrule{1-14}
GPT-4o \citep{openai2024gpt4o} & 62.30 & 64.35 & 59.21 & 62.38 & 64.75 & 60.00 & 58.88 & 59.50 & 58.25 & 48.46 & 50.90 & 44.67 & 84.33 \\
Gemini-3-Flash \citeyearpar{googledeepmind2025gemini3flash} & 86.91 & 83.48 & 92.11 & 90.75 & 91.00 & 90.50 & 88.38 & 87.50 & 89.25 & 62.22 & 64.59 & 58.53 & 85.33 \\
\midrule
\multicolumn{14}{c}{\textbf{\textit{Open-Source Model}}} \\
\cmidrule{1-14}
Qwen2.5-VL-7B \citep{bai2025qwen25vl} & 76.44 & 79.13 & 72.37 & 67.38 & 78.75 & 56.00 & 63.75 & 73.75 & 53.75 & 43.82 & 48.16 & 37.07 & 65.67 \\
\quad + text-only SFT & 80.63 & 85.22 & 73.68 & 68.38 & 78.28 & 58.50 & 62.88 & 73.50 & 52.25 & 49.92 & 52.87 & 45.33 & 69.67 \\
LLaVA-OneVision \citep{li2024llavaonevision} & 69.63 & 73.91 & 63.16 & 65.12 & 75.75 & 54.50 & 57.50 & 65.75 & 49.25 & 45.49 & 49.44 & 39.33 & 77.00 \\
InternVL3-8B \citep{zhu2025internvl3} & 78.53 & 77.39 & 80.26 & 70.13 & 83.25 & 57.00 & 64.50 & 76.00 & 53.00 & 51.38 & 54.15 & 47.07 & 78.33 \\
Vision-R1 \citep{huang2026visionr1} & 80.10 & 80.87 & 78.95 & 73.38 & 83.75 & 63.00 & 70.12 & 79.25 & 61.00 & 50.81 & 54.32 & 45.33 & 73.00 \\
\midrule
\multicolumn{14}{c}{\textbf{\textit{Thinking-with-Images Model}}} \\
\cmidrule{1-14}
ZoomEye \citep{shen2025zoomeye} & 83.77 & 87.83 & 77.63 & 70.00 & 83.00 & 57.00 & 65.63 & 78.50 & 52.75 & 47.06 & 50.04 & 42.40 & 71.33 \\
Thyme \citep{zhang2025thyme} & 82.20 & 83.48 & 80.26 & 78.00 & 87.75 & 68.25 & 73.00 & 82.50 & 63.50 & 52.27 & 56.12 & 46.27 & 73.67 \\
DeepEyes \citep{zheng2026deepeyes} & 83.25 & 83.48 & 82.89 & 73.00 & 85.75 & 60.25 & 66.63 & 79.50 & 53.75 & 53.93 & 57.23 & 48.80 & 73.67 \\
\midrule
\multicolumn{14}{c}{\textbf{\textit{Visual Latent Reasoning Model}}} \\
\cmidrule{1-14}
LVR \citep{li2025lvr} & 81.15 & 82.61 & 78.95 & 69.25 & 81.50 & 57.00 & 64.88 & 73.25 & 56.50 & 50.03 & 54.66 & 42.80 & 72.00 \\
SkiLa \citep{tong2025skila} & 80.10 & 80.87 & 78.95 & 71.63 & 85.75 & 57.50 & 67.63 & 80.75 & 54.50 & 51.54 & 55.26 & 45.73 & 73.33 \\
Monet \citep{wang2025monet} & 81.15 & 80.00 & 82.89 & 71.25 & \textbf{87.00} & 55.50 & 67.25 & 79.25 & 55.25 & 50.18 & 53.98 & 44.27 & 68.00 \\
HyLaR \citep{cheng2026hylar} & 81.15 & 80.87 & 81.58 & 73.13 & 86.50 & 59.75 & 67.25 & 82.50 & 52.00 & 47.33 & 51.07 & 40.67 & 73.67 \\
\textbf{GLaQ-SFT (Ours)} & 82.72 & 81.74 & \textbf{84.21} & 71.13 & 84.75 & 57.50 & 65.63 & 80.00 & 54.00 & 51.80 & 55.18 & 47.53 & 70.00 \\
\textbf{GLaQ-7B (Ours)} & \textbf{85.34} & \textbf{86.09} & \textbf{84.21} & \textbf{73.37} & 85.50 & \textbf{61.25} & \textbf{70.37} & \textbf{83.75} & \textbf{57.00} & \textbf{53.31} & \textbf{56.97} & \textbf{47.60} & \textbf{75.33} \\
\textit{$\Delta$ (pts.) vs. Qwen2.5-VL-7B} & +8.90 & +6.96 & +11.84 & +5.99 & +6.75 & +5.25 & +6.62 & +10.00 & +3.25 & +9.49 & +8.81 & +10.53 & +9.66 \\
\bottomrule
\end{tabular*}
}
\caption{Main results on V$^\ast$, HRBench-4K, HRBench-8K, MME-RealWorld-Lite, and MMVP. All results are obtained from our reproductions and evaluated with VLMEvalKit. Scores follow official benchmark metrics; higher is better. $\Delta$ reports absolute gains over Qwen2.5-VL-7B. The visual latent model with the best performance is highlighted in \textbf{bold}.}
\label{tab:main_results}
\end{table*}

\begin{table*}[t]
\centering
\begin{minipage}[t]{0.48\textwidth}
\vspace{0pt}
\centering
{\small
\setlength{\tabcolsep}{1.2pt}
\begin{tabular}{lccccc}
\toprule
Model & V$^\ast$ & HR-4K & HR-8K & MME-Lite & MMVP \\
\midrule
\textbf{GLaQ-7B (full)} & \textbf{85.34} & \textbf{73.37} & \textbf{70.37} & \textbf{53.31} & \textbf{75.33} \\
GLaQ-SFT (w/o RL) & 82.72 & 71.13 & 65.63 & 51.80 & 70.00 \\
\quad + text-only GRPO & 84.82 & 72.00 & 69.38 & 52.42 & 74.00 \\
w/o EMA teacher & 80.15 & 69.38 & 62.75 & 47.99 & 67.67 \\
w/o LQFormer & 78.54 & 67.13 & 62.38 & 46.87 & 66.67 \\
Text-only SFT & 80.63 & 68.38 & 62.88 & 49.92 & 69.67 \\
\bottomrule
\end{tabular}
}
\caption{Component ablations of GLaQ. All SFT variants use the same training data, and all RL variants start from the same GLaQ-SFT checkpoint.}
\label{tab:component_ablation}
\end{minipage}
\hfill%
\begin{minipage}[t]{0.48\textwidth}
\vspace{0pt}
\centering
{\small
\setlength{\tabcolsep}{2.0pt}
\begin{tabular}{lccccc}
\toprule
$\lambda_{\mathrm{latent}}$ & V$^\ast$ & HR-4K & HR-8K & MME-Lite & MMVP \\
\midrule
$0$   & 80.10 & 68.75 & 63.00 & 47.84 & 70.67 \\
$0.1$ & 81.15 & 69.75 & 65.88 & 48.98 & 71.33 \\
$0.3$ & \textbf{82.20} & 70.13 & \textbf{66.38} & \textbf{49.19} & \textbf{73.67} \\
$0.5$ & 80.58 & \textbf{70.63} & 65.75 & 48.72 & 72.67 \\
$1.0$ & 78.01 & 68.75 & 64.75 & 48.67 & 72.00 \\
\bottomrule
\end{tabular}
}
\caption{Sensitivity to the latent-alignment weight $\lambda_{\mathrm{latent}}$. Results are obtained under the Stage 1 SFT setting with all other settings fixed.}
\label{tab:lambda_stage1_ablation}
\end{minipage}
\end{table*}

\subsection{Reinforcement Learning with DePO}
\label{sec:latent-rl}

During reinforcement learning, GLaQ produces categorical text tokens and continuous grounded queries, whereas standard token-level policy optimization directly handles only the former. We therefore adopt DePO from HyLaR \citep{cheng2026hylar}, treating post-LQFormer queries $\mat{Z}_Q$ as continuous actions and decoded tokens as categorical actions. The fixed $K$-slot layout aligns corresponding latent actions across the old, current, and reference policies.

Each rollout is partitioned into text positions $\mcal{P}^{\mathrm{text}}$ and latent positions $\mcal{P}^{\mathrm{latent}}$, with $\mcal{P}=\mcal{P}^{\mathrm{text}}\cup\mcal{P}^{\mathrm{latent}}$ and each query block contributing $K$ latent positions. Let $\Theta=(\theta,\phi,\mat{Q})$ denote the trainable policy parameters. DePO then optimizes the following joint text--latent objective:
\begin{equation}
\begin{aligned}
\mathcal{J}_{\mathrm{DePO}}(\Theta)
&=
\mathcal{J}_{\mathrm{text}}
\left(
\Theta \mid \mcal{P}^{\mathrm{text}};
\epsilon_{\ell}^{\mathrm{text}},
\epsilon_{h}^{\mathrm{text}}
\right)\\
&\quad+
\alpha\,
\mathcal{J}_{\mathrm{latent}}
\left(
\Theta \mid \mcal{P}^{\mathrm{latent}};
\epsilon_{\ell}^{\mathrm{latent}},
\epsilon_{h}^{\mathrm{latent}}
\right)\\
&\quad-
\beta_{\mathrm{text}}
D_{\mathrm{KL}}^{\mathrm{text}}
-
\beta_{\mathrm{latent}}
D_{\mathrm{KL}}^{\mathrm{latent}} .
\end{aligned}
\end{equation}
At text positions, $\mathcal{J}_{\mathrm{text}}$ uses categorical likelihood ratios. At latent positions, the old policy records normalized rollout queries, while the current and frozen reference policies recompute their directions through GLaQ's two-pass pathway. DePO applies von Mises--Fisher density ratios, a scaled-cosine KL term, and separate clipping ranges, making grounded retrieval directly optimizable under task rewards.

For each prompt $(\mat{I},\vect{x})$, the old policy samples $G$ responses and computes GRPO-style group-normalized advantages \citep{shao2024deepseekmath}. We use the weighted reward $R(\vect{o}_i)=(1-\lambda_{\mathrm{fmt}})R_{\mathrm{acc}}(\vect{o}_i)+\lambda_{\mathrm{fmt}}R_{\mathrm{fmt}}(\vect{o}_i)$, where $\lambda_{\mathrm{fmt}}=0.1$ and the two binary terms check answer correctness and \texttt{\textbackslash boxed\{\}} formatting, respectively. We add no separate latent-reasoning reward, keeping optimization focused on task success.

\section{Experiments}

\label{sec:experiments}

\subsection{Experimental Settings}

\label{sec:experimental-settings}

\paragraph{Training Data.}
GLaQ uses two supervised datasets and one reinforcement-learning dataset. Stage 1 retains 80K Visual-CoT samples \citep{shao2024visualcot} where Qwen2.5-VL-7B \citep{bai2025qwen25vl} fails on the original image but succeeds with an auxiliary reasoning view. The paired views provide full-image and localized evidence for student--teacher latent-query alignment. Stage 2 uses 30K Zebra-CoT visual-search examples \citep{li2025zebracot}, whose interleaved trajectories train grounded queries for multi-step evidence seeking. The reinforcement-learning dataset comprises 10K samples from DeepEyes and Thyme \citep{zheng2026deepeyes,zhang2025thyme} after removing duplicates, subjective or open-ended questions, and samples without unambiguous, extractable answers to support verifiable answer-level rewards.

\paragraph{Benchmarks.}
Following prior visual latent reasoning studies \citep{tong2025skila,cheng2026hylar,wang2025monet}, we evaluate on five complementary benchmarks: V$^\ast$ for fine-grained attribute and spatial grounding \citep{wu2024vstar}; HRBench-4K/8K for high-resolution perception and cross-instance reasoning \citep{wang2025hrbench}; MME-RealWorld-Lite for complex real-world perception and reasoning \citep{zhang2025mmerealworld}; and MMVP for subtle visual-difference recognition \citep{tong2024mmvp}.

\paragraph{Baselines.}
To compare GLaQ with general-purpose MLLMs and visual reasoning methods, we consider four baseline groups: (1) \textbf{Proprietary Model}, including GPT-4o and Gemini-3-Flash \citep{openai2024gpt4o,googledeepmind2025gemini3flash}; (2) \textbf{Open-Source Model}, including Qwen2.5-VL-7B, LLaVA-OneVision, InternVL3, and Vision-R1 \citep{bai2025qwen25vl,li2024llavaonevision,zhu2025internvl3,huang2026visionr1}; (3) \textbf{Thinking-with-Images Model}, including ZoomEye, Thyme, and DeepEyes \citep{shen2025zoomeye,zhang2025thyme,zheng2026deepeyes}; and (4) \textbf{Visual Latent Reasoning Model}, including LVR, SkiLa, HyLaR, and Monet \citep{li2025lvr,tong2025skila,cheng2026hylar,wang2025monet}.

\paragraph{Implementation Details.}
All GLaQ variants use a two-layer LQFormer, $K=16$ grounded query slots, and a Qwen2.5-VL-7B-Instruct backbone \citep{bai2025qwen25vl}. All reported benchmark results are obtained from our reproductions using the same VLMEvalKit evaluation settings \citep{duan2024vlmevalkit}. GLaQ is evaluated on the original benchmark images without auxiliary localized views and uses deterministic decoding.

\subsection{Main Results}

\label{sec:main-results}

\paragraph{Overall performance.}
Table~\ref{tab:main_results} shows that GLaQ-7B improves its Qwen2.5-VL-7B base model by 5.99--9.66\% and achieves the highest overall score among the compared visual latent methods on all five benchmarks. More importantly, the results reveal a resolution-dependent contrast on HRBench. As the image resolution increases from 4K to 8K, the improvement on single-instance questions grows from 6.75\% to 10.00\%, whereas the improvement on cross-instance questions decreases from 5.25\% to 3.25\%. This pattern is consistent with grounded queries being particularly effective at recovering localized evidence that becomes difficult to preserve in high-resolution visual representations, while integrating evidence across multiple instances remains more challenging. The gains on MME-RealWorld-Lite and MMVP further indicate that this advantage is not confined to the visual-search trajectories used during supervised training, but extends to complex real-world scenes and subtle visual distinctions. GLaQ also remains competitive with thinking-with-images methods while using only the original images, suggesting that internal query-to-image grounding can provide an effective alternative to external visual operations.

\paragraph{Training-stage improvements.}
The progression across training stages reveals complementary roles for supervised grounding and policy optimization. GLaQ-SFT consistently outperforms text-only SFT, showing that supervision routed through the grounded-query pathway contributes beyond textual adaptation alone. However, the supervised model leads prior visual latent methods on only two of the five benchmarks, suggesting that latent alignment establishes a useful grounding prior but leaves room for task-level adaptation. DePO produces its largest additional gains on HRBench-8K and MMVP and extends GLaQ's category-leading performance from two benchmarks to all five. This concentration of gains on visually demanding settings is consistent with task-level feedback helping the model not only acquire fine-grained evidence, but also use the grounded representations more effectively for final prediction.

\subsection{Component Ablations}
\label{sec:ablations}

Table~\ref{tab:component_ablation} shows that the supervised gains arise from exploiting intermediate visual supervision through explicit query-to-image grounding and teacher-guided alignment, rather than from inserting continuous query states alone. We construct the text-only SFT baseline from the same supervised examples by masking the auxiliary reasoning views and applying standard next-token prediction to the remaining textual sequences. The original problem images are retained, while the grounded-query blocks and latent-alignment objective are removed. GLaQ-SFT consistently outperforms this baseline, indicating that the auxiliary views become more useful when their localized evidence is transferred to full-image grounded queries. Removing the EMA teacher eliminates most of this advantage, suggesting that LQFormer benefits from localized targets that specify what evidence should be extracted from the full image. More strikingly, bypassing LQFormer causes performance to fall below text-only SFT on all five benchmarks, even though the first-pass query states are still reused as soft prompts. This controlled contrast indicates that continuous query states are insufficient by themselves; their benefit depends on renewed, context-conditioned access to the source visual tokens. Together, these ablations indicate complementary roles: LQFormer enables direct query-to-image interaction, while self-distillation guides the grounded queries toward task-relevant evidence.

The policy ablation further separates the general benefit of reinforcement learning from optimizing grounded queries. Starting from the same GLaQ-SFT model with matched training data, rewards, and rollout settings, text-only GRPO already yields substantial gains, confirming the importance of text-token policy optimization. DePO further improves this matched variant by 0.52--1.37\% on every benchmark, showing that grounded queries are not merely passive representations but optimizable latent actions that benefit from task-level feedback. DePO therefore complements supervised alignment by refining how grounded visual evidence supports final prediction.

\subsection{Hyperparameter Sensitivity}
\label{sec:hyperparameter-sensitivity}

We examine GLaQ's sensitivity to the latent-alignment weight $\lambda_{\mathrm{latent}}$ and the number of grounded query slots $K$, which control supervision strength and query capacity, respectively. Both studies use the Stage 1 SFT setting and vary one hyperparameter at a time while holding the remaining settings fixed for a controlled comparison.

\paragraph{Latent-alignment weight.}
Table~\ref{tab:lambda_stage1_ablation} shows that alignment strength, rather than its presence alone, determines how well latent supervision complements answer learning. The moderate setting $\lambda_{\mathrm{latent}}=0.3$ is the most consistent, ranking first on four of the five benchmarks; increasing the weight to $0.5$ benefits only HRBench-4K, while $1.0$ is weaker than $0.3$ throughout. The isolated HRBench-4K gain may reflect a favorable balance at this resolution: stronger ROI alignment helps preserve fine-grained details, while the model can still integrate sufficient global context. This non-monotonic trend indicates that latent query self-distillation functions best as an auxiliary grounding regularizer. Stronger emphasis can overconstrain the student toward ROI-conditioned teacher representations and compete with task-adaptive next-token prediction. Overall, the optimal balance is mildly task-dependent, but $0.3$ provides the most robust operating point.

\begin{table}[!t]
\centering
\setlength{\tabcolsep}{2.5pt}
\begin{tabular}{lccccc}
\toprule
$K$ & V$^\ast$ & HR-4K & HR-8K & MME-Lite & MMVP \\
\midrule
$8$  & 78.01 & 68.75 & 65.00 & 47.99 & 73.33 \\
$16$ & \textbf{82.20} & \textbf{70.13} & \textbf{66.38} & \textbf{49.19} & \textbf{73.67} \\
$32$ & 78.53 & 68.50 & 63.75 & 49.09 & 70.00 \\
\bottomrule
\end{tabular}
\caption{Sensitivity to the number of grounded query slots $K$. Results are obtained under the Stage 1 SFT setting with $\lambda_{\mathrm{latent}}=0.3$ and all other settings fixed.}
\label{tab:k_ablation}
\end{table}

\paragraph{Grounded query slots.}
Table~\ref{tab:k_ablation} reveals a non-monotonic relationship between query capacity and performance. Although $K=16$ performs best on all five benchmarks, $K=32$ outperforms $K=8$ on V$^\ast$ and MME-RealWorld-Lite but falls behind it on both HRBench settings and MMVP. This reversal reflects a trade-off between evidence coverage and representation redundancy. Increasing $K$ provides more capacity to retrieve distinct cues, but the slots attend to the same visual-token pool and are not explicitly constrained to specialize; with $K=32$, overlapping or weakly relevant queries may therefore dilute the grounded signal when jointly reinjected, providing a plausible explanation for its lower performance than $K=8$ on some tasks. The consistent advantage of $K=16$ suggests a more stable balance between coverage and redundancy. Since $K$ widens a single visual lookup rather than adding new observations or reasoning steps, this trend should not be interpreted as a test-time scaling effect.

\subsection{End-to-End Generation Efficiency}
\label{sec:accuracy-throughput}

\begin{figure}[t]
\centering
\includegraphics[width=\columnwidth]{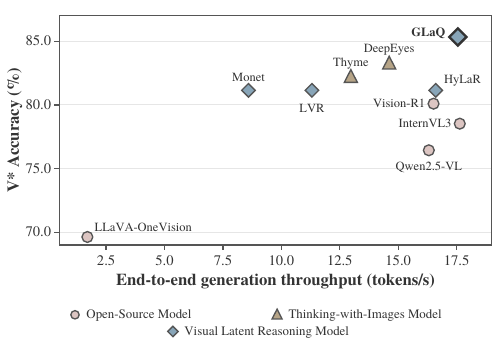}
\caption{Accuracy--throughput trade-off on V$^\ast$. All methods are evaluated on the same 191 examples under the same setup.}
\label{fig:vstar-throughput}
\end{figure}

We measure end-to-end generation throughput as the total number of generated tokens divided by the total wall-clock time, including all visual and language model operations. Figure~\ref{fig:vstar-throughput} shows that GLaQ advances the accuracy--efficiency frontier across all three model families. Relative to its Qwen2.5-VL-7B backbone, DeepEyes, and HyLaR, GLaQ improves V$^\ast$ accuracy by 8.90, 2.09, and 4.19 points, respectively, while increasing throughput by 7.62\%, 20.11\%, and 5.70\%. The gain over the backbone suggests that context-conditioned queries improve evidence recovery through renewed access to source visual tokens, while the fixed query block keeps this additional grounding computation bounded. Because each query block has fixed capacity, its grounding overhead remains bounded, unlike variable-length latent rollouts whose cost grows with additional latent states. The larger throughput advantage over DeepEyes further highlights the benefit of performing grounding internally: GLaQ preserves full-image context without external visual operations or repeated image processing. Meanwhile, its simultaneous gains over HyLaR suggest that grounding queries directly in the source image can mitigate the evidence drift and redundancy of sequential latent construction, while avoiding variable-length autoregressive latent rollouts. Together, these results show that GLaQ improves visual evidence access through bounded internal grounding, yielding gains in both accuracy and efficiency.

\section{Conclusion}
We introduced GLaQ, a grounded latent-query framework that reformulates continuous visual reasoning as context-conditioned visual lookup. GLaQ contextualizes, grounds, and reinjects fixed query blocks, preserving direct access to source visual representations without autoregressive latent-state construction. Latent query self-distillation transfers localized supervision to the full-image pathway, while DePO optimizes textual and grounded-query actions. Across five benchmarks, GLaQ consistently improves its Qwen2.5-VL-7B backbone and ranks first among the compared visual latent reasoning methods. Across the ablation and efficiency analyses, improvements consistently arise when latent queries regain direct access to source tokens and remain coordinated within a fixed-capacity block. These results support explicit source-image grounding as an effective alternative to autoregressive latent construction. Our evaluation is limited to one 7B backbone, fixed query capacity, and auxiliary localized-view supervision. Future work will examine model scaling, adaptive query allocation, and grounding objectives that reduce reliance on localized views.

\bibliography{aaai2027}

\end{document}